\documentclass[sigconf, screen]{acmart}
\AtBeginDocument{%
  }

\setcopyright{acmlicensed}
\copyrightyear{2026}
\acmYear{2026}
\acmDOI{XXXXXXX.XXXXXXX}
\acmConference[MM 26]{ACM International Conference on Multimedia}{10–-14 November, 2026}{Rio de Janeiro, Brazil}

\acmISBN{978-1-4503-XXXX-X/2018/06}

\usepackage{multirow}
\usepackage{wrapfig}
\usepackage{graphicx}
\usepackage{booktabs}
\usepackage{enumitem}
\usepackage{overpic}    % overpic 环境

\usepackage{pifont}
\newcommand{\xmark}{\ding{55}}  % 定义 \xmark 命令
\newcommand{\cmark}{\ding{51}}  % 定义对号

\definecolor{ForestGreen}{RGB}{34,139,34}
\definecolor{mycolor}{RGB}{255, 99, 71}
\definecolor{DarkGreen}{RGB}{0,100,0}
\definecolor{MyGreen}{RGB}{84,202,26}
\renewcommand\footnotetextcopyrightpermission[1]{} %remove copyright
\begin{document}

\title{Benchmarking and Improving GUI Agents in High-Dynamic Environments}

\author{Enqi Liu$^{1,2}$, Liyuan Pan$^{1,3}$, Zhi Gao$^{1,2}$, Yan Yang$^{4}$, Chenrui Shi$^{1,2}$, Yang Liu$^{2}$, Jingrong Wu$^{2}$, Qing Li$^{2}$}

\email{enqi.liu@bit.edu.cn}

\affiliation{%
  \institution{
  $^{1}$ Beijing Institute of Technology, Beijing, China\\
  $^{2}$ Beijing Institute for General Artificial Intelligence, Beijing, China\\
  $^{3}$ Yangtze Delta Region Academy of Beijing Institute of Technology, Jiaxing, China\\
  $^{4}$ Australian National University, Canberra, Australia
  }
  \country{}
}

\renewcommand{\shortauthors}{Trovato et al.}

\begin{abstract}
Recent advancements in Graphical User Interface (GUI) agents have predominantly focused on training paradigms like supervised fine-tuning (SFT) and reinforcement learning (RL). However, the challenge of high-dynamic GUI environments remains largely underexplored. Existing agents typically rely on a single screenshot after each action for decision-making, leading to a partially observable (or even unobservable) Markov decision process, where the key GUI state including important information for actions is often inadequately captured. To systematically explore this challenge, we introduce DynamicGUIBench, a comprehensive online GUI benchmark spanning ten applications and diverse interaction scenarios characterized by important interface changes between actions. Furthermore, we present DynamicUI, an agent designed for dynamic interfaces, which takes screen-recording videos of the interaction process as input and consists of three components: a dynamic perceiver, a refinement strategy, and a reflection. Specifically, the dynamic perceiver clusters frames of the GUI video, generates captions for the centroids, and iteratively selects the most informative frames as the salient dynamic context. Considering that there may be inconsistencies and noise between the selected frames and the textual context of the agent, the refinement strategy employs an action-conditioned filtering to refine thoughts to mitigate thought-action inconsistency and redundancy. Based on the refined agent trajectories, the reflection module provides effective and accurate guidance for further actions. Experiments on DynamicGUIBench demonstrate that DynamicUI significantly improves the performance in dynamic GUI environments, while maintaining competitive performance on other public benchmarks.
\end{abstract}

% \begin{CCSXML}
% <ccs2012>
%    <concept>
%        <concept_id>10010147.10010178</concept_id>
%        <concept_desc>Computing methodologies~Artificial intelligence</concept_desc>
%        <concept_significance>500</concept_significance>
%        </concept>
%  </ccs2012>
% \end{CCSXML}

% \ccsdesc[500]{Computing methodologies~Artificial intelligence}

% \keywords{GUI Agent, VLM-based Agent, Real-World Tasks}
\maketitle
\pagestyle{empty}
\section{Introduction}

\begin{figure}[t]
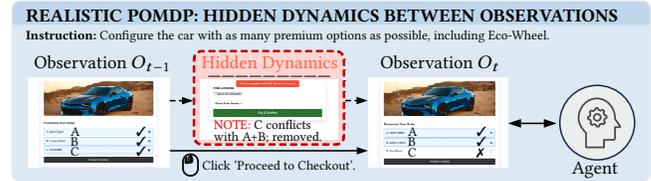

    \centering
    \begin{overpic}[width=0.48\textwidth]
    {The_Name_of_the_Title_Is_Hope__1_/samples/pictures/fig1_finals.pdf}
    \put(2.5,25.3){\footnotesize \textbf{\textcolor{black}{REALISTIC POMDP: HIDDEN DYNAMICS BETWEEN OBSERVATIONS}}}
    \put(2.5,22.5){\tiny \text{\textcolor{black}{\textbf{Instruction:} Configure the car with as many premium options as possible, including Eco-Wheel.}}}
    \put(3.8,18){\footnotesize \text{\textcolor{black}{Observation $O_{t-1}$}}}
    \put(9.5,7.5){\tiny \text{\textcolor{black}{A}}}
    \put(19.7,7.5){\tiny \text{\textcolor{black}{\cmark}}}
    \put(9.5,6.0){\tiny \text{\textcolor{black}{B}}}
    \put(19.7,6.0){\tiny \text{\textcolor{black}{\cmark}}}
    \put(9.5,4.4){\tiny \text{\textcolor{black}{C}}}
    \put(19.7,4.4){\tiny \text{\textcolor{black}{\cmark}}}
    \put(31.8,8.6){\tiny \text{\textcolor{black}{\textcolor{red}{NOTE:} C conflicts}}}
    \put(31.8,6.8){\tiny \text{\textcolor{black}{ with A+B; removed.}}}

    \put(30,2.3){\tiny \text{\textcolor{black}{Click `Proceed to Checkout'.}}}

    \put(62,7.55){\tiny \text{\textcolor{black}{A}}}
    \put(72.7,7.55){\tiny \text{\textcolor{black}{\cmark}}}
    \put(62,5.9){\tiny \text{\textcolor{black}{B}}}
    \put(72.7,5.9){\tiny \text{\textcolor{black}{\cmark}}}
    \put(62,4.25){\tiny \text{\textcolor{black}{C}}}
    \put(72.7,4.25){\tiny \text{\textcolor{black}{\xmark}}}

    \put(57.7,18){\footnotesize \text{\textcolor{black}{Observation $O_{t}$}}}
    \put(87.5,1.6){\footnotesize \text{\textcolor{black}{Agent}}}
    \put(29.9,18){\footnotesize \text{\color{mycolor}{Hidden Dynamics}}}
    \put(30.8,16){\footnotesize \text{\textcolor{black}{---------------------------}}}
    \end{overpic}
    \caption{Illustration of how hidden dynamics make task-relevant state information only partially observable to current GUI agents. In this example, a configuration mismatch prompt appears between two observations, so the current observation $O_t$ does not fully preserve the temporal information required by the task, even when considering the previous observation $O_{t-1}$.}
    \label{fig:MDP}
\end{figure}

Leveraging Vision-Language Models (VLMs) to develop GUI agents has emerged as a promising frontier for automating complex tasks, such as file processing and information retrieval, across mobile and desktop platforms~\cite{huang2025spiritsight, yang2025aria, ye2025mobile, liulabel, zhao2025learning, lei2025gui}. Given the domain gap between natural environments and GUI screenshots, research has predominantly focused on developing strong learning algorithms~\cite{yao2022react, wang2025mobilea3gent, yuan2025enhancing}, constructing large-scale, high-quality datasets~\cite{gou2024navigating, bai2025qwen3} or architecting sophisticated systems~\cite{gonzalez2025unreasonable, yang2025gta1} to improve GUI comprehension. While these advancements are undoubtedly valuable, most existing agents~\cite{zhang2026tongui, songcoact, team2026kimi} rely on a singular screenshot captured after each action for decision-making. This paradigm overlooks the temporal nature of GUI interactions, failing to account for high-dynamic environments where visual changes of the interfaces (e.g., transient notifications, streaming animations, or page scrolling) occur between discrete action steps while these changes may not be fully observable from a single screenshot and can lead to the loss of critical task-relevant information, as shown in Fig.~\ref{fig:MDP}. Consequently, a single-frame observation is insufficient to fully capture the evolving GUI state, reducing the interaction to a partially observable Markov decision process (POMDP) and stripping the GUI agents of essential context for informed planning.

Although recent studies such as D-GARA~\cite{chen2025d} and GUI-Robust~\cite{yang2025gui} have begun to consider `dynamic' GUI tasks, they mainly model dynamics as observable anomalies or perturbations, such as unexpected advertisements or warning dialogs. Such settings still implicitly follow a fully observable MDP assumption, where the post-action screenshot is treated as sufficient to represent the environment state. This view overlooks the hidden interstitial dynamics that may occur between two consecutive observations. In real-world GUI environments, however, crucial task-relevant events may emerge, evolve, or disappear before the next screenshot is captured, making the underlying interaction process inherently partially observable. As a result, important evidence for instruction understanding and action selection can be lost under sparse temporal sampling.

To bridge this gap, we introduce DynamicGUIBench, a comprehensive online GUI benchmark with 149 tasks across ten applications, explicitly designed to evaluate GUI agents under hidden interstitial dynamics. In contrast to traditional benchmarks that primarily assume static post-action interfaces, DynamicGUIBench focuses on four representative categories of dynamic challenges, namely interruptive UI states, ephemeral reference, dynamic list selection, and content-triggered interaction. These categories capture a broad range of realistic cases in which the true environment state cannot be reliably recovered from a single screenshot alone. By explicitly modeling such partially observable interaction processes, DynamicGUIBench provides a more rigorous and realistic testbed for evaluating robust GUI agents.

We further propose DynamicUI, a novel agent framework that takes the full interaction screen recording video as input and consists of three modules: a dynamic perceiver, a refinement strategy, and a reflection module. The dynamic perceiver condenses the raw video into informative dynamic context by clustering frames and iteratively selecting salient centroids based on generated captions and confidence scores. The refinement strategy further mitigates inconsistencies between the agent’s intended actions and resulting executable behaviors. Finally, the reflection module leverages the refined trajectory to provide high-level guidance for subsequent actions. Extensive experiments on DynamicGUIBench and other public benchmarks show that DynamicUI is highly competitive with existing state-of-the-art methods, particularly in dynamic GUI scenarios, while maintaining strong general GUI understanding.

Our main contributions are as follows.
\begin{itemize}
\setlength\itemsep{0.0em}
    \item We introduce DynamicGUIBench, an online benchmark spanning ten applications and diverse dynamic GUI scenarios, where substantial interface changes occur between actions, which make the interaction process partially observable for existing agents.
    \item We propose DynamicUI, a GUI agent for rapidly changing interfaces, built on a dynamic perceiver, a trajectory refinement strategy, and a reflection module.
    \item Extensive experiments on DynamicGUIBench and the OSWorld benchmark demonstrate that DynamicUI achieves strong performance in dynamic environments while remaining competitive in static settings.
\end{itemize}
\section{Related Work}
\noindent{\bf GUI Benchmark.}~Existing GUI benchmarks can be broadly grouped by the type of tasks they target. One major line focuses on short-horizon, step-wise action prediction~\cite{cheng2024seeclick, xie2024osworld}, where the goal is to predict the next click, type, or grounded action from a single screenshot or a short observation history. These benchmarks mainly evaluate perception, grounding, and immediate decision-making ability at the current step. A second line emphasizes long-horizon goal completion~\cite{deng2023mind2web, longlonghorizonui, qin2025osgym}, requiring agents to perform multi-step reasoning, planning, backtracking, and recovery in order to accomplish an instruction end-to-end in realistic environments. A third line studies cross-application workflows~\cite{sun2025gui, gao2025chain, xu2024aguvis}, where agents must coordinate actions across multiple apps, windows, or webpages, and transfer intermediate information between heterogeneous interfaces. In addition, several recent benchmarks explicitly stress robustness under anomalies or disruptions~\cite{chen2025d, yang2025gui, rawles2024androidworld}, such as pop-up dialogs, permission prompts, interruptions, or environment perturbations, in order to better approximate practical deployment settings.

However, despite differences in task scope and evaluation protocol, existing benchmarks still assume a fully observable MDP, where each screenshot is treated as a sufficient representation of the current state and transitions are assumed not to contain critical latent events. Even when anomalies are considered, they are typically modeled as directly observable interruptions, rather than missing or partially observed intermediate states.

In contrast, our benchmark is designed for POMDP-style GUI tasks, where substantial interface changes and latent events may occur between two observations. As a result, a single frame observation can be incomplete or even misleading with respect to the true state. This setting better reflects real-world desktop and mobile interaction, where agents must reason over dynamic context, partial evidence, and temporally sparse but task-critical events.

\noindent{\bf GUI Agent.}
~Recent progress in GUI agents has been driven by advances in both agentic models and agent frameworks, mainly under supervised fine-tuning (SFT) and reinforcement learning (RL) paradigms. SFT-based methods fine-tune large vision--language models on GUI datasets to improve UI grounding, action prediction, and instruction following, as exemplified by ShowUI~\cite{lin2025showui}, CogAgent~\cite{hong2024cogagent}, HATS~\cite{shao2026hats}, SeeClick~\cite{cheng2024seeclick}, and SimpAgent~\cite{chen2025less}. RL-based approaches, such as UI-AGILE~\cite{lian2025ui} and GUI-R1~\cite{luo2025gui}, improve decision-making through interaction feedback and reward-driven optimization. Beyond single-stage training, multi-stage or modular pipelines such as GTA1~\cite{yang2025gta1}, Aguvis~\cite{xu2024aguvis}, and related planner--executor systems introduce reasoning, grounding, and verification modules to enhance execution in complex environments.

At the same time, another important direction explores stronger foundation backbones for GUI control, including general-purpose VLMs and native GUI-action models such as Qwen-VL~\cite{bai2025qwen3}, OpenCUA~\cite{wang2025opencua}, GUI-Owl~\cite{xu2026mobile}, UIPro~\cite{li2025uipro}, and UI-TARS~\cite{qin2025ui}. These models show that scaling multimodal pretraining and instruction tuning can substantially improve action grounding and long-horizon control. However, such gains mainly stem from stronger perception and action generation, rather than explicit modeling of hidden states or missing observations in dynamic interfaces.

Most existing methods improve execution through expert imitation, reward shaping, stronger backbones, or modular design, rather than explicitly enhancing dynamic state tracking under partial observability. As a result, they remain limited in POMDP-style GUI tasks, where critical intermediate states may be latent or unobserved. In contrast, we propose a training-free framework that can be readily integrated into diverse VLM-based agents to strengthen dynamic context modeling in GUI interaction.

% \vspace{-3em}
\section{DynamicGUIBench}

\subsection{Data Collection and Annotation}
To capture real-world dynamics in desktop environments, such as unexpected pop-up dialogs, system prompts, and user-driven interface changes, we develop a structured data construction pipeline, as illustrated in Fig.~\ref{fig:3}. The pipeline consists of five stages.

First, annotators propose task ideas grounded in everyday desktop workflows. Second, a VLM expands them into candidate instructions, which are refined for clarity, executability, and consistency with the benchmark design. Third, annotators create automation configurations for task initialization and dynamic event triggering. Fourth, they define task-specific evaluation functions and instantiate the corresponding evaluators to form complete task specifications. Finally, all tasks are double-checked by multiple annotators for correctness and overall quality.

\begin{figure}[t]
    \centering
    \begin{overpic}[width=0.48\textwidth]
    {The_Name_of_the_Title_Is_Hope__1_/samples/pictures/3_cropped.pdf}
       \put(12,33){\footnotesize \textbf{\textcolor{black}{Human}}}
       \put(12,26.8){\footnotesize \textbf{\textcolor{black}{Script}}}
       \put(12,20){\footnotesize \textbf{\textcolor{black}{VLM}}}

       \put(30,18.5){\footnotesize \textbf{\textcolor{black}{Inspiration}}}
       \put(37,26){\scriptsize \textit{\textbf{\textcolor{black}{Propose}}}}
       \put(52.5,26){\scriptsize \textit{\textbf{\textcolor{black}{Extend}}}}
       \put(86,25){\scriptsize \textit{\textbf{\textcolor{black}{Build}}}}
       \put(12.5,5){\scriptsize \textit{\textbf{\textcolor{black}{Judge}}}}
       \put(43,5){\scriptsize \textit{\textbf{\textcolor{black}{Generate}}}}
       \put(79.5,5){\scriptsize \textit{\textbf{\textcolor{black}{Check}}}}
       \put(53,34.0){\footnotesize \textbf{\textcolor{black}{Instruction}}}
       \put(43,35.5){\scriptsize \textit{\textbf{\textcolor{black}{Check}}}}

       \put(67.5,18.5){\footnotesize \textbf{\textcolor{black}{Instruction}}}
       \put(0.8,13.5){\footnotesize \textbf{\textcolor{black}{Instruction}}}
       \put(57.4,6){\footnotesize \textbf{\textcolor{black}{EFunc.}}}

       \put(47.1,15.2){\footnotesize \textbf{\textcolor{black}{01}}}
       \put(79.9,15.2){\footnotesize \textbf{\textcolor{black}{02}}}
       \put(14.3,1.8){\footnotesize \textbf{\textcolor{black}{03}}}
       \put(47.1,1.8){\footnotesize \textbf{\textcolor{black}{04}}}
       \put(79.9,1.8){\footnotesize \textbf{\textcolor{black}{05}}}

       \put(4.9,3.5){\footnotesize \textbf{\textcolor{red}{\xmark}}}
       \put(27.5,3.5){\footnotesize \textbf{\textcolor{ForestGreen}{\cmark}}}
       \put(72,3.5){\footnotesize \textbf{\textcolor{red}{\xmark}}}
       \put(97.4,3.5){\footnotesize \textbf{\textcolor{ForestGreen}{\cmark}}}
       \footnotesize
    \end{overpic}
    \caption{Pipeline of data construction. EFunc. represents evaluator and reward function.
    }
    \label{fig:3}
    
\end{figure}

To improve reproducibility and reduce variance caused by network conditions and environment updates, we provide fully self-contained offline assets for Chrome-based and multi-app tasks, where complete HTML files are constructed to ensure deterministic rendering and interaction behavior. These design choices make DynamicGUIBench more reproducible, extensible, and practical for standardized evaluation.

\begin{table}[t]
    \centering
    \small
    \caption{Comparison of existing GUI benchmarks and our benchmark in terms of application diversity, task scale, evaluation setting (Mode), anomaly presence (An.), and dynamic tasks (Dyn.). \cmark\ and \xmark\ denote the presence and absence of anomaly or dynamic tasks, respectively.}
    \setlength{\tabcolsep}{2.0pt}
    \begin{tabular}{lccccccc}
        \toprule
        \textbf{Benchmark} & \textbf{Venue} & \textbf{Apps/Web} & \textbf{Tasks} 
        & \textbf{Mode} & \textbf{An.} & \textbf{Dyn.} \\
        \midrule
        Mind2Web~\cite{deng2023mind2web} & NeurIPS'23 & 137  & 2,350 & Offline & \xmark & \xmark \\
        OSWorld~\cite{xie2024osworld} & NeurIPS'24 & 10  & 369 & Online & \xmark & \xmark \\
        GUI Odyssey~\cite{lu2025guiodyssey} & ICCV'25 & 201  & 7,735 & Offline & \xmark & \xmark \\
        WorldGUI~\cite{hengyuan2025worldgui} & arXiv'25 & 10  & 315 & Offline & \xmark & \xmark \\
        Android World~\cite{rawles2405androidworld} & ICLR'25 & 20  &116 & Offline & \xmark & \xmark \\
        OSWorld-G~\cite{xie2025scaling} & NeurIPS'25 &  10 &564 & Online & \xmark & \xmark \\
        GUI-Robust~\cite{yang2025gui} & NeurIPS'25 & 392 & 5,318 & Offline & \cmark & \xmark \\
        D-GARA~\cite{chen2025d} & AAAI'26 & 7 & 152 & Online & \cmark & \xmark \\
        \midrule
        Ours & 2026 & 10 & 149 & Online & \cmark & \cmark \\
        \bottomrule
    \end{tabular}
    \label{tab:1}
\end{table}

\subsection{Benchmark Analysis}

Tab.~\ref{tab:1} compares DynamicGUIBench with representative GUI agent benchmarks. Existing benchmarks largely fall into two groups: (i) offline datasets constructed from static traces, such as Mind2Web~\cite{deng2023mind2web}, GUI Odyssey~\cite{lu2025guiodyssey}, GUI-ReWalk~\cite{lin2025gui}, GUI Knowledge Bench~\cite{shi2025gui}, and GUI-Robust~\cite{yang2025gui}, and (ii) online interactive environments, such as OSWorld~\cite{xie2024osworld}, Android World~\cite{rawles2405androidworld}, and D-GARA~\cite{chen2025d}. 
While GUI-Robust and D-GARA explicitly introduce anomalous situations, the task state in these benchmarks remains fully observable at each step, such that these scenarios can still be handled by existing methods with partial-observability modeling and do not fundamentally depart from a fully observable Markov process. 
In contrast, DynamicGUIBench introduces an online, dynamic environment where latent interstitial states make the task partially observable and sometimes non-Markovian, thereby filling a key gap in existing benchmarks by enabling systematic evaluation of GUI agents under dynamic context shifts and incomplete observability.

\begin{table}[t]
    \centering
     \caption{Key statistics in DynamicGUIBench.}
        \label{tab:stats}
        % \vspace{0.4em}
        \small
        \setlength{\tabcolsep}{10.0pt}
        \begin{tabular}{lccc}
            \toprule
            \textbf{Statistic} & \textbf{Number} & \textbf{Statistic} & \textbf{Number}\\
            \midrule
            Chrome            & 38 (25.5\%)& Multi apps  & 33 (22.4\%) \\
            Gimp              & 10 (6.8\%)& Os    & 7 (4.7\%) \\
            Libreoffice$_{ \text{calc} }$  & 14 (9.5\%)& Thunderbrid & 9 (6.1\%) \\
            Libreoffice$_{ \text{impress} }$ & 7 (4.7\%)& VLC   & 5 (3.4\%) \\
            Libreoffice$_{ \text{writer} }$ & 14 (9.5\%)& VS Code & 12 (8.1\%) \\
            \midrule
            Feasible & 146 (98\%)&  Infeasible & 3 (2.0\%) \\
            \bottomrule
        \end{tabular}
    \label{tab:stats}
\end{table}

Tab.~\ref{tab:stats} summarizes the benchmark statistics. DynamicGUIBench contains 149 tasks spanning 10 application domains, covering a broad range of realistic desktop interaction scenarios. Chrome and multi-application tasks account for the largest shares, while 146 tasks are feasible and 3 are infeasible, enabling evaluation of both execution capability and infeasibility awareness. Built on top of OSWorld, the benchmark supports direct interactive evaluation while preserving realistic task dynamics. More details on the benchmark construction are provided in the supplementary material.

\begin{figure*}[t]
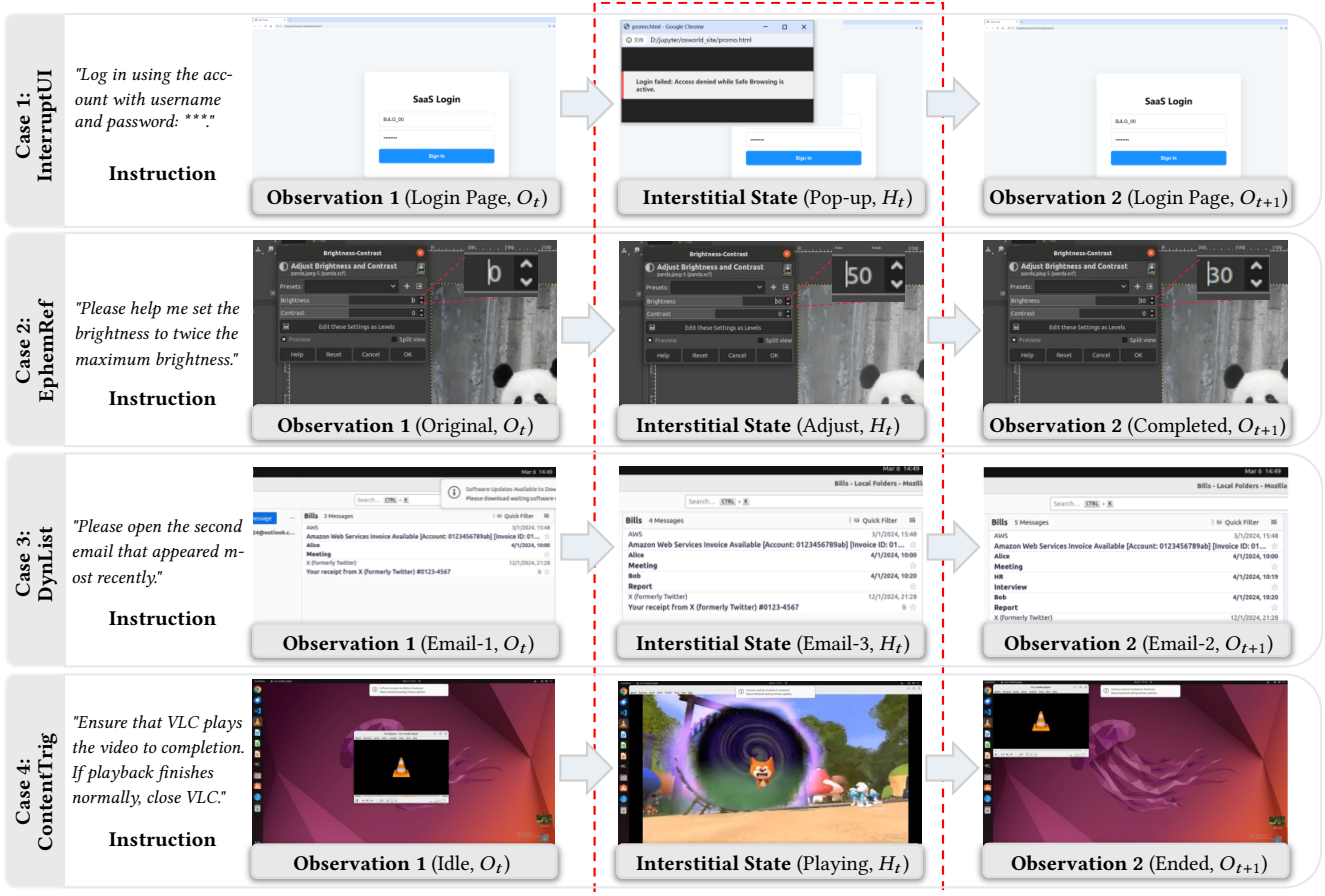

    \centering
    \begin{overpic}[width=1\textwidth]
    {The_Name_of_the_Title_Is_Hope__1_/samples/pictures/cases4_0330_cropped.pdf}
        \put(1.2,54){
\rotatebox{90}{
\small\textbf{\shortstack[c]{Case 1:\\
InterruptUI}}
}}
\put(1.2,37.6){
\rotatebox{90}{
\small\textbf{\shortstack[c]{Case 2: \\ EphemRef}}
}}
\put(1.2,22.6){
\rotatebox{90}{
\small\textbf{\shortstack[c]{Case 3: \\DynList}}
}}
\put(1.2,4){
\rotatebox{90}{
\small\textbf{\shortstack[c]{Case 4: \\ContentTrig}}
}}
 \put(20.4,52.8){\small{\textcolor{black}{\shortstack{\textbf{Observation 1} (Login Page, $O_t$)}}}}
 \put(48.5,52.8){\small{\textcolor{black}{\shortstack{\textbf{Interstitial State} (Pop-up, $H_t$)}}}}
 \put(74.4,52.8){\small{\textcolor{black}{\shortstack{\textbf{Observation 2} (Login Page, $O_{t+1}$)}}}}
 \put(21.2,35.8){\small{\textcolor{black}{\shortstack{\textbf{Observation 1} (Original, $O_t$)}}}}
 \put(48,35.8){\small{\textcolor{black}{\shortstack{\textbf{Interstitial State} (Adjust, $H_t$)}}}}
 \put(74.4,35.8){\small{\textcolor{black}{\shortstack{\textbf{Observation 2} (Completed, $O_{t+1}$)}}}}
\put(22.4,3){\small{\textcolor{black}{\shortstack{\textbf{Observation 1} (Idle, $O_t$)}}}}
 \put(48,3){\small{\textcolor{black}{\shortstack{\textbf{Interstitial State} (Playing, $H_t$)}}}}
 \put(76,3){\small{\textcolor{black}{\shortstack{\textbf{Observation 2} (Ended, $O_{t+1}$)}}}}
\put(21.6,19.5){\small{\textcolor{black}{\shortstack{\textbf{Observation 1} (Email-1, $O_t$)}}}}
 \put(48,19.5){\small{\textcolor{black}{\shortstack{\textbf{Interstitial State} (Email-3, $H_t$)}}}}
 \put(75.5,19.5){\small{\textcolor{black}{\shortstack{\textbf{Observation 2} (Email-2, $O_{t+1}$)}}}}

 \put(6,40.8){{\footnotesize\selectfont
    \shortstack[l]{%
    \it "Please help me set the  \\
    \it brightness to twice the \\
    \it maximum brightness."%
    }}}
\put(8.6,37.8){\small{\textcolor{black}{\shortstack{\textbf{Instruction}}}}}
 \put(5.8,8.0){{\footnotesize\selectfont
    \shortstack[l]{%
    \it "Ensure that VLC plays   \\
    \it the video to completion.   \\
    \it If playback finishes \\
    \it normally, close VLC."
    }}}
\put(8.6,4.8 ){\small{\textcolor{black}{\shortstack{\textbf{Instruction}}}}}

 \put(5.8,24.5){{\footnotesize\selectfont
    \shortstack[l]{%
    \it "Please open the second   \\
    \it  email that appeared m-   \\
    \it ost recently."
    }}}
\put(8.6,21.3 ){\small{\textcolor{black}{\shortstack{\textbf{Instruction}}}}}

\put(6,58.6){{\footnotesize\selectfont
    \shortstack[l]{%
    \it "Log in using the acc-  \\
    \it  ount with username     \\
    \it  and password: ***."
    }}}
\put(8.6,54.6){\small{\textcolor{black}{\shortstack{\textbf{Instruction}}}}}
    \end{overpic}
    \caption{Representative cases from DynamicGUIBench. Each example illustrates a hidden interstitial state $H_t$ (marked by the \textcolor{red}{red dashed box}) arising between two consecutive observations, $O_t$ and $O_{t+1}$. 
    Case 1: \text{InterruptUI}, interrupted by a blocking dialog.
Case 2: \text{EphemRef}, dependent on short-lived interaction history.
Case 3: \text{DynList}, dependent on relative ordering in a changing list.
Case 4: \text{ContentTrig}, triggered by temporally localized semantic content.}

    \label{fig:examples}
\end{figure*}

\subsection{POMDP Design}
To systematically characterize dynamic interstitial states in desktop environments, we organize DynamicGUIBench under a POMDP taxonomy with four categories:
\begin{itemize}
\setlength\itemsep{0.0em}

\item \textbf{Interruptive UI states} (InterruptUI). Sudden inserted interface states, such as warnings, security prompts, login failures, or configuration errors, that interrupt the original workflow and must be handled before the task can proceed.

\item \textbf{Ephemeral reference} (EphemRef). Tasks that rely on short-lived interaction history, such as recent selections, previous steps, or last-used settings, which may not be recoverable from the current screenshot alone.

\item \textbf{Dynamic list selection} (DynList). Candidate sets such as inboxes, feeds, and search results may change over time, requiring selection based on relative attributes (e.g., recency, rank, or price) rather than fixed positions.

\item \textbf{Content-triggered interaction} (ContentTrig). Actions are triggered by semantic content in text, images, videos, or subtitles, where the decisive evidence may be brief, implicit, or not fully captured by a single static screenshot.
\end{itemize}

\begin{figure}[t]
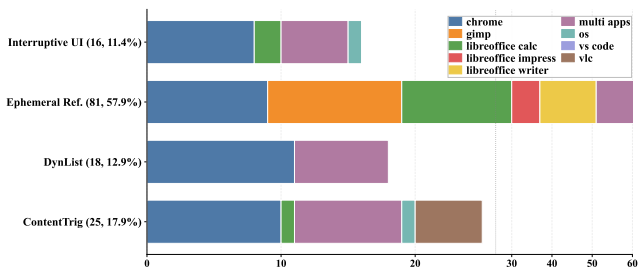

  \centering
  % \vspace{-0.8cm}
  \begin{overpic}[width=0.48\textwidth]
    {The_Name_of_the_Title_Is_Hope__1_/samples/pictures/dynamicguibench_stacked_bar_final_0330.pdf}
  \end{overpic}
  \caption{Task distribution in DynamicGUIBench across four POMDP categories, with colors indicating applications.}
  \label{fig:statistics}
\end{figure}

Fig.~\ref{fig:examples} presents one representative example for each of the four categories, showing that the environment may evolve between two recorded observations while the decisive intermediate evidence is not fully reflected in either endpoint screenshot. These cases illustrate that the central challenge is not merely the presence of visible anomalies, but the omission of temporally localized yet task-critical information under sparse observation. Such scenarios arise naturally in everyday web and desktop interactions and constitute common sources of failure for GUI agents in realistic settings. As shown in Fig.~\ref{fig:statistics}, DynamicGUIBench covers all four categories with diverse distributions across application domains. More importantly, these categories induce different forms of partial observability, including interrupted workflows, short-lived references, evolving candidate sets, and transient semantic cues, which are difficult to capture faithfully with static web-style benchmarks.

\section{Method}
\subsection{Formulation}
\label{formulation}
We formulate GUI task automation as a sequential decision-making problem, where the agent controller is parameterized by a vision-language-action (VLA) model $M_{\theta}$ (such as Qwen3-vl-8B~\cite{bai2025qwen3} and UITARS-1.5-7B~\cite{qin2025ui}).
\begin{figure*}[t]
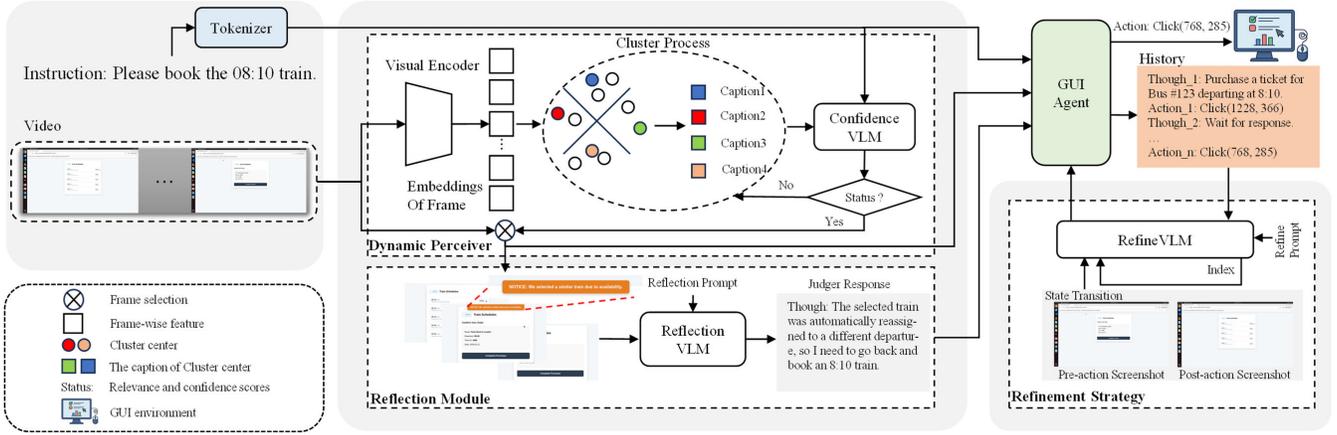

    \centering
    \begin{overpic}[width=1\textwidth]
    {The_Name_of_the_Title_Is_Hope__1_/samples/pictures/frameworks_0402.pdf}
    \end{overpic}
    \caption{The overall architecture of DynamicUI. The system comprises three collaborative components: (1) Dynamic Perceiver utilizes a Visual Encoder and a Cluster Process to capture key dynamic visual information. (2) Reflection Module analyzes task-specific screenshots and generate corrective thoughts when inconsistencies arise. (3) Refinement Strategy prunes redundant information and improves the accuracy of action execution. This closed-loop mechanism ensures precise GUI navigation by aligning visual perception with iterative logical reasoning through state transitions.
    }
    \label{fig:2}
\end{figure*}
At each time step $i$, the agent observes the inference video $v_i$, which records the visual trajectory over the previous $n$ steps, together with the historical thoughts and actions ($r_{i-n}$, $a_{i-n}$, $\dots$, $r_{i-1}$, $a_{i-1}$), the output of the reflection module $f_{i-1}$, and the task instruction $q$. The agent generates a new thought $r_i$ and an executable action $a_i$ from the action space, such as clicking on a specific UI element, entering text, or scrolling through the interface. Executing the action $a_i$ results in a new environment state $v_{i+1}$. The interaction loop continues iteratively, with the agent observing the updated environment, generating decisions, and executing actions until a termination condition is satisfied (e.g., task completion or failure) or a predefined maximum number of steps is reached,
\begin{equation}
r_i^\star, a_i^\star = \arg \max_{r_i, a_i} 
M_{\theta}(r_i, a_i \mid q, v_i, r_{i-n}, a_{i-n}, \dots, a_{i-1}, f_{i-1}).
\end{equation}
The whole framework is shown in Fig.~\ref{fig:2}.

\subsection{Dynamic Perceiver}
Specifically, at time step $i$, we record all $m$ frames from the first operation to the $i$-th operation as a video sequence $v_i=\{o_1,o_2,\ldots,o_m\}$, which provides dynamic context for the current decision. In contrast, relying solely on the current observation $o_m$, as in many prior methods, incorrectly reduces the underlying POMDP to a fully observable MDP. This simplification may overlook key hidden dynamics across actions, leading to incomplete instruction understanding and unsuccessful task execution.

\begin{figure}[t]
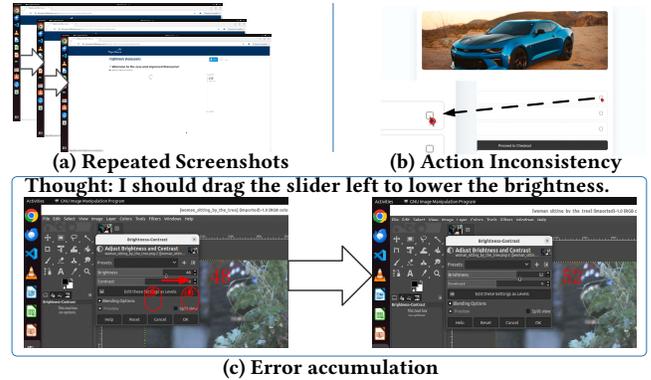

  \centering
  \begin{overpic}[width=0.48\textwidth]
    {The_Name_of_the_Title_Is_Hope__1_/samples/pictures/fig6_finals.pdf}
  \put(7,32){\small \textbf{\textcolor{black}{(a) Repeated Screenshots}}}
  \put(59.5,32){\small \textbf{\textcolor{black}{(b) Action Inconsistency}}}
  \put(2.7,28.3){\small \textbf{\textcolor{black}{Thought: I should drag the slider left to lower the brightness.}}}
  \put(33.5,0){\small \textbf{\textcolor{black}{(c) Error accumulation}}}
  \end{overpic}
  \caption{Three failure modes of traditional GUI agents.}
  \label{fig:limitation}
\end{figure}

To preserve informative dynamic context while suppressing redundant observations, we use $v_i$ as the input representation of the GUI agent. At each step, every frame is encoded by a visual encoder $E_v$ to obtain frame-level features $\mathbf{z}_t=E_v(o_t)$ for $t\in\{1,\ldots,m\}$. We then cluster these features to capture meaningful dynamic variations across the interaction trajectory while consolidating near-duplicate frames, thereby mitigating the repeated-screenshot issue shown in Fig.~\ref{fig:limitation} (a). Formally,
\begin{equation}
\{\mathcal{C}_c\}_{c=1}^{C},\{\boldsymbol{\mu}_c\}_{c=1}^{C} = \mathrm{Cluster}(\{\mathbf{z}_t\}_{t=1}^{m},C),
\end{equation}
where $C$ is initialized to 3. Each cluster roughly represents a distinct stage of the task. We then generate a caption for each cluster center as $\hat{y}_c=M_s(q_s\mid \boldsymbol{\mu}_c)$ using a VLM $M_s$, and feed both $(\boldsymbol{\mu}_c,\hat{y}_c)$ together with the current instruction $q$ into another VLM $M_{\mathrm{conf}}$,
\begin{equation}
(r_c,\kappa_c)=M_{\mathrm{conf}}(q,v_i,\boldsymbol{\mu}_c,\hat{y}_c), c\in \{1,\ldots,C\},
\end{equation}
where $r_c \in [0, 3]$ denotes the relevance score between the $c$-th cluster center and the current instruction, with a higher score indicating that this cluster is more likely to provide useful task-relevant evidence, and $\kappa_c \in [0, 100]$ denotes the confidence score, measuring the confidence of model in its own relevance judgment.

\begin{figure*}[t]
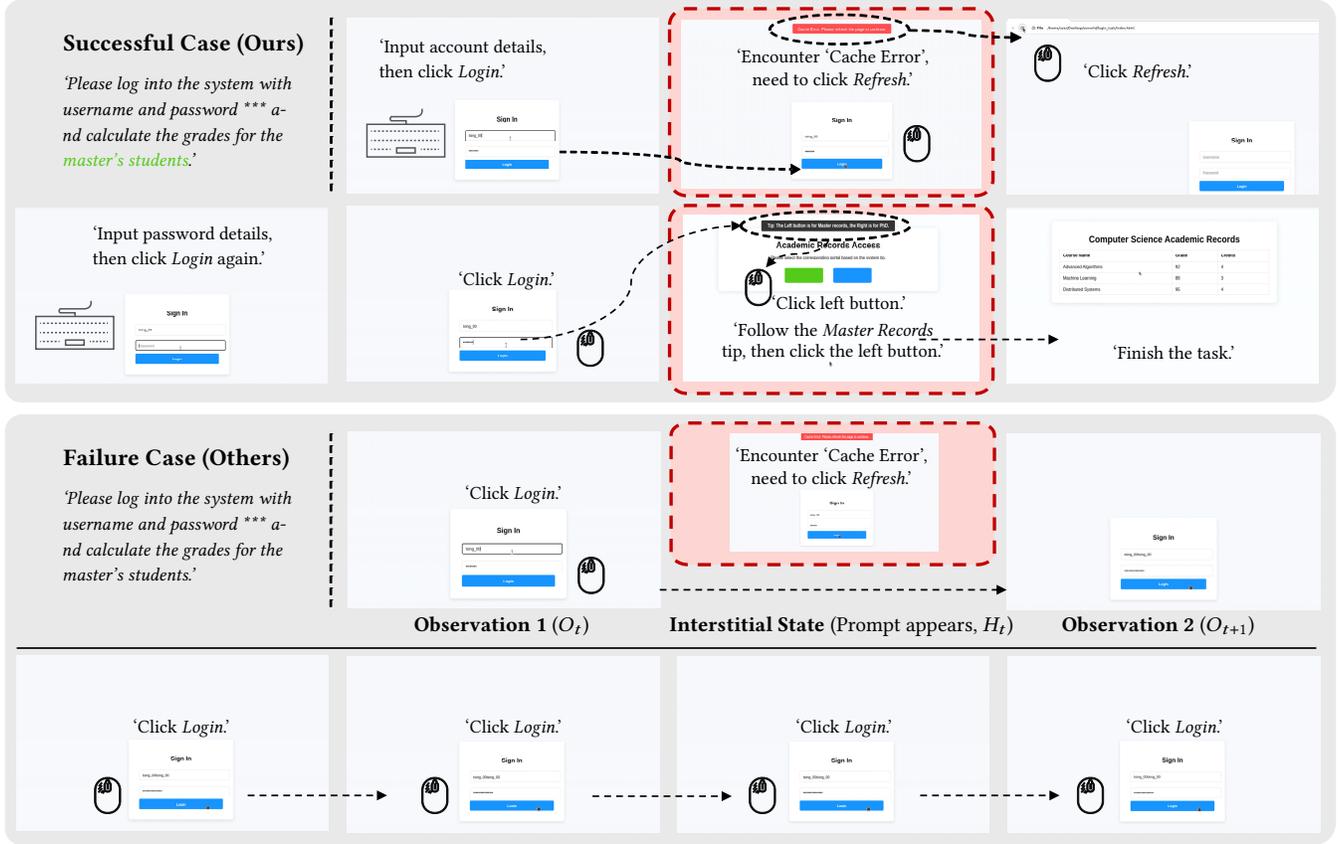

    \centering
    \begin{overpic}[width=1\textwidth]
    {The_Name_of_the_Title_Is_Hope__1_/samples/pictures/fig7_final_0402.pdf}
    \put(4.6,59.8){\normalsize{\textcolor{black}{\shortstack{\textbf{Successful Case (Ours)}}}}}
    \put(4.5,51.2){{\footnotesize\selectfont
    \shortstack[l]{%
    \it `Please log into the system with   \\
    \it  username and password *** a-   \\
    \it nd calculate the grades for the \\
    \it \textcolor{MyGreen}{master's students}.'
    }}}
    \put(4.6,28.8){\normalsize{\textcolor{black}{\shortstack{\textbf{Failure Case (Others)}}}}}
    \put(4.5,20.2){{\footnotesize\selectfont
    \shortstack[l]{%
    \it `Please log into the system with   \\
    \it  username and password *** a-   \\
    \it nd calculate the grades for the \\
    \it master's students.'
    }}}
    \put(28.2,57.8){{\footnotesize\selectfont
    \shortstack[l]{%
    `Input account details,   \\
    then click \textit{Login}.'
    }}}
    \put(54.9,57.2){{\footnotesize\selectfont
    \shortstack[c]{%
    `Encounter `Cache Error',   \\
     need to click \textit{Refresh}.'
    }}}
    \put(80.8,57.8){{\footnotesize\selectfont
    \shortstack[l]{%
    `Click \textit{Refresh}.' 
    }}}
    \put(6.8,43.8){{\footnotesize\selectfont
    \shortstack[l]{%
    `Input password details,    \\
     then click \textit{Login} again.'
    }}}
    \put(34.1,42.3){{\footnotesize\selectfont
    \shortstack[l]{%
    `Click \textit{Login}.'
    }}}
    \put(57.5,40.5){{\footnotesize\selectfont
    \shortstack[c]{%
    `Click left button.'
    }}}
    \put(53.8,36.9){{\footnotesize\selectfont
    \shortstack[c]{%
    `Follow the \textit{Master Records}  \\
     tip, then click the left button.'
    }}}
    \put(83,36.8){{\footnotesize\selectfont
    \shortstack[l]{%
    `Finish the task.'
    }}}
    \put(34.6,26.3){{\footnotesize\selectfont
    \shortstack[l]{%
    `Click \textit{Login}.'
    }}}
    \put(30.8,16.4){\small{\textcolor{black}{\shortstack{\textbf{Observation 1} ($O_t$)}}}}
     \put(49.9,16.4){\small{\textcolor{black}{\shortstack{\textbf{Interstitial State} (Prompt appears, $H_t$)}}}}
     \put(54.8,27.4){{\footnotesize\selectfont
    \shortstack[c]{%
    `Encounter `Cache Error',   \\
     need to click \textit{Refresh}.'
    }}}
    \put(79.2,16.4){\small{\textcolor{black}{\shortstack{\textbf{Observation 2} ($O_{t+1}$)}}}}

    \put(9.8,8.8){{\footnotesize\selectfont
    \shortstack[l]{%
    `Click \textit{Login}.'
    }}}
    \put(34.6,8.8){{\footnotesize\selectfont
    \shortstack[l]{%
    `Click \textit{Login}.'
    }}}
    \put(59.3,8.8){{\footnotesize\selectfont
    \shortstack[l]{%
    `Click \textit{Login}.'
    }}}
    \put(84,8.8){{\footnotesize\selectfont
    \shortstack[l]{%
    `Click \textit{Login}.'
    }}}
    
    \end{overpic}
    \caption{A representative comparison between DynamicUI (top) and traditional GUI agents (bottom) on DynamicGUIBench. DynamicUI captures the interstitial state $H_t$, identifies the task-critical prompt \textit{Cache Error, then click `Refresh'}, and successfully completes the task. In contrast, traditional GUI agents miss this transient cue and get stuck repeatedly clicking the `Login' button. The red boxes highlight task-critical hidden dynamics.}
    \label{fig:success_failure}
\end{figure*}

If the scores do not meet predefined thresholds, we iteratively refine the clustering by increasing the number of clusters, 
\begin{equation}
C \leftarrow 2C \quad \text{if} \quad \max_{c} r_c < \tau_r \ \text{or}\ \max_{c} \kappa_c < \tau_{\kappa},
\end{equation}
until convergence or a predefined maximum number of clusters is reached, with $\tau_r = 3$ and $\tau_{\kappa} = 80$. We then select the qualified cluster centers as key frames, denoted by $\{\tilde{o}_i\}$ where $i \in \mathcal{I}$ indexes their positions in the original video, and provide them to the GUI agent as visual context.

\subsection{Refinement Strategy}
After the Dynamic Perceiver selects the instruction-relevant frames $\tilde{o}_i$ and the GUI agent predicts the next action $a_i$, the generated thought may be inconsistent with the executed action or the resulting outcome. As illustrated in Fig.~\ref{fig:limitation} (b), although the agent predicts the intended action, the executed click does not satisfy the expectation described in the thought, indicating that the click coordinates need further refinement. To address this issue, we propose an action-conditioned refinement strategy to improve the consistency and reliability of subsequent decision-making.

Specifically, we introduce a VLM $M_F$ to jointly refine the retained thought--action pairs based on the executed action, the neighboring screenshots, and the visual prompt on the resulting screenshot, which together reveal whether the intended operation has been correctly carried out. This step corrects cases where the original thought appears plausible but the action is incorrectly executed,
\begin{equation}
\tilde{r}_t, \tilde{a}_t
= M_F\!\left( r_t, a_t\,\middle|\, q_f,\ a_t,\ o_{t-1},\ o_t \right),
\end{equation}
where $\tilde{r}_t$ and $\tilde{a}_t$ denote the refined thought and refined action for step $t$, respectively, and $q_f$ represents the system prompt of $M_F$.

\begin{table*}[t]
    \centering
    \small
    \caption{ Comparison with state-of-the-art methods on the DynamicGUIBench. Abbreviations: Th.\ (Thunderbird), Multi.\ (Multi-Apps), Vs.\ (Visual Studio Code), Imp.\ (Impress), Wri.\ (Writer). The best scores are in bold.}
    \hspace{-0cm}
    \setlength{\tabcolsep}{9.4pt}
    \begin{tabular}{lccccccccccc|c}
        \toprule
        \multirow{2}{*}{\textbf{Agent Model}} 
        & \multirow{2}{*}{\textbf{Step}} 
        & \multicolumn{11}{c}{\textbf{Acc (\%)}} \\
        \cmidrule(lr){3-13}
        &
        & Chrome & Gimp & Calc & Imp. & Wri. & Multi. & Os & Th. & Vlc & Vs. & Avg. \\
        \midrule
        o3~\cite{Openai_o3}  & 15 
        & 7.9  & 10.0 & 7.1 & 0.0 &7.1  &6.1  &0.0  &0.0  & 20.0 &8.3  & 6.7 \\
        Qwen3-vl-4B~\cite{bai2025qwen3}  & 15 
        &18.4 &10.0  &7.1  &0.0  & 7.1 & 9.1 &14.3  &0.0  &40.0  & 0.0 &  10.7\\
        Qwen3-vl-8B~\cite{bai2025qwen3}  & 15 
        & 21.1 &20.0  &{14.3}  &0.0  &14.3  &9.1  &14.3  &0.0 &40.0  &16.7  &14.8 \\
        o3~\cite{Openai_o3} & 50 
        & 10.5  & 0.0 & {14.3} &0.0  &7.1  & 6.1 & {42.9} &0.0  & 20.0 &0.0  &8.7  \\
        UITARS-1.5-7B~\cite{qin2025ui}  & 50 
        & 13.2 &10.0  &7.1  &0.0  &7.1  &9.1  & 0.0 & 0.0 &20.0  & 0.0 & 8.1 \\
        
        Qwen3-vl-4B~\cite{bai2025qwen3}  & 50 
        & 13.2  &10.0  &7.1  &0.0  &7.1  &9.1  &28.6  &0.0  &{60.0}  &0.0  &10.7  \\
        Qwen3-vl-8B~\cite{bai2025qwen3}  & 50 
        & 26.3  & 22.0 & {14.3} & 0.0 &7.1  & 7.5 & 14.3 &0.0  &{60.0}  & 8.3 & 15.1 \\
        doubao-1-5-0717\cite{guo2025seed1}  & 50 
        & 5.3  &0.0  & 7.1 &0.0  &0.0  &0.0  &0.0  & 0.0 &0.0  &0.0  &2.8  \\
        Seed1.8-VL~\cite{seedseed1}  & 50 
        & 10.5  &10.0  & 0.0 &{14.3}  & 7.1 & 3.0 & 0.0 & 0.0 & 0.0 &{16.7}  & 6.7 \\
        EvoCUA-8B~\cite{xue2026evocua} & 50 
        & 18.4  & {33.0} &0.0  &0.0  &7.1  &4.5  &28.6  &0.0  & 20.0 &{16.7}  &11.7  \\

        \midrule
        Ours w/ Qwen3-vl-8B  & 15 
        &29.0  &10.0  &{14.3}  &0.0  &7.1  &{12.2}  &14.3  &{11.1}  &0.0  &{16.7}  &\textbf{15.5}\\

         Ours w/ Qwen3-vl-8B  & 50 
        &{36.8}  &30.0  &14.3  & 14.3 &7.1  & 9.1 &28.6  &44.4  &40.0  &8.3 &\textbf{22.1} \\
        \bottomrule
    \end{tabular}
    \label{tab:main}
\end{table*}

\subsection{Reflection Module}
As shown in Fig.~\ref{fig:limitation} (c), the agent’s reasoning can drift away from the actual task objective during multi-step interaction. In this example, the thought suggests dragging the brightness slider to the left, yet the executed action moves it to the right. Such incorrect actions are recorded in the interaction history and may further bias subsequent reasoning, leading to compounding failures. To alleviate this issue, we introduce an auxiliary VLM, denoted as $M_R$, as a reflection module to better evaluate task progress and provide corrective guidance for subsequent actions.

Given the reflection prompt $q_r$, selected key screenshots $\tilde{o}_{i-1}$, the textual interaction history, and the corresponding actions, the reflection module produces an auxiliary textual feedback $f_i$, which is then used to guide the GUI agent in generating subsequent actions and improve next-step action prediction,

\begin{equation}
f_i = M_{R}(q_r, \tilde{o}_{i-1}, r_{i-n}, a_{i-n}, \cdots, r_{i-1}, a_{i-1}).
\end{equation}

Please refer to the supplementary material for more details.

\section{Experiments}
\subsection{Implementation Details}

We compare two representative paradigms of GUI agents: proprietary API-based agents (e.g., doubao-1-5-0717~\cite{guo2025seed1}, o3~\cite{Openai_o3}, and Seed1.8-VL~\cite{seedseed1}) and open-source agentic models (e.g., Qwen3-vl~\cite{bai2025qwen3}, UITARS-1.5~\cite{qin2025ui}, and EvoCUA~\cite{xue2026evocua}). Unless otherwise specified, the maximum number of interaction steps is set to 50 for all methods.

\subsection{Experimental Results}
To systematically evaluate dynamic reasoning and interaction capabilities, we benchmark DynamicUI against a diverse set of state-of-the-art agent models on DynamicGUIBench. As shown in Tab.~\ref{tab:main}, both open-weight models (e.g., Qwen3-VL-8B) and proprietary API-based agents (e.g., o3) achieve rather limited performance, highlighting the difficulty of dynamic GUI tasks for current agents.

Among all baselines, Qwen3-VL-8B achieves the strongest overall performance with an average accuracy of 15.1\% under the 50-step setting, while most other methods remain below this level by a large margin. In particular, several models completely fail on certain applications, and some categories such as Thunderbird remain especially challenging: both o3 and Qwen3-VL-4B obtain 0.0\% accuracy on this domain. These results indicate that existing agents struggle to robustly handle dynamically changing interface states, temporal updates, and interstitial events.

In contrast, DynamicUI consistently outperforms all baselines. Under the 50-step setting, our method achieves the best average accuracy of 22.1\%, surpassing the strongest baseline, Qwen3-VL-8B~\cite{bai2025qwen3}, by 7.0 absolute percentage points. Moreover, DynamicUI achieves the best or second-best performance across all 10 application domains, including substantial gains on Chrome (36.8\%), and Thunderbird (44.4\%),. Notably, DynamicUI is the only method that achieves non-zero accuracy on Thunderbird, highlighting its advantage in handling highly dynamic email-based tasks.

\begin{table}[t]
    \centering
    \caption{Ablation on the DynamicGUIBench.}
    \small
    \setlength{\tabcolsep}{12.5pt}
        \begin{tabular}{ccc|c}
                \toprule
                \shortstack{\textbf{Dynamic}\\\textbf{Perceiver}} &
\shortstack{\textbf{Reflection}\\\textbf{Module}} &
\shortstack{\textbf{Refinement}\\\textbf{Strategy}} &
\shortstack{\textbf{Acc(\%)}}  \\
\midrule                
                - & - & - & 15.1 \\
                \cmark & - &  & 17.4 \\
                \cmark & - & \cmark & 17.4  \\
                \cmark & \cmark & - &20.8  \\
                \cmark& \cmark & \cmark &\textbf{22.1}   \\
                \bottomrule
            \end{tabular}
    \label{tab:aba}
\end{table}

\begin{table*}[t]
    \centering
    \small
    \caption{Comparison with state-of-the-art methods on the OSWorld~\cite{xie2024osworld} benchmark. We use `*' to denote the results evaluated by us (which will be updated if improved evaluation scripts become available).}
    \hspace{-0cm}
    \setlength{\tabcolsep}{9.35pt}
    \begin{tabular}{lccccccccccc|c}
        \toprule
        \multirow{2}{*}{\textbf{Agent Model}} 
        & \multirow{2}{*}{\textbf{Step}} 
        & \multicolumn{11}{c}{\textbf{Acc (\%)}} \\
        \cmidrule(lr){3-13}
        &
        & Chrome & Gimp & Calc & Imp. & Wri. & Multi. & Os & Th. & Vlc & Vs. & Avg. \\
        \midrule
        o3~\cite{Openai_o3}  & 50 
        & 21.7  &38.5  & 8.5 &4.3  & 21.7 &11.8  & 37.5 & 20.0 &17.6  & 21.7 & 17.2 \\
        Opencua-a3b~\cite{wang2025opencua}  & 50 
        &21.6  &53.9  &2.1 &23.3 & 34.8 & 6.5 &37.5  &6.7 & 11.8 &43.5  &19.9  \\
        UITARS-72b-dpo~\cite{qin2025ui}  & 50 
        &33.2  &61.5  &12.8 &25.5 & 43.5 & 6.7 &33.3  &33.3  & 23.5 &47.8  &25.8  \\
        UITARS-1.5-7B~\cite{qin2025ui}  & 50 
        &32.8   &53.9  &8.5  & 39.3 & 41.3 &8.6  &25.6  &46.7  & 24.4 &52.2  & 27.2 \\
        % Qwen3-vl-4B*~\cite{bai2025qwen3}  & 50 
        % & -  &-  &- & - & - & - &-  &-  & - &-  &26.2  \\
        Qwen3-vl-4B*~\cite{bai2025qwen3}  & 50 
        & 30.4  &21.7  &12.8 & 21.8 & 39.1 & 5.2 &33.3  &46.7  & 23.0 &13.0  &19.8  \\
        Opencua-7B~\cite{wang2025opencua}  & 50 
        &38.6  &43.6  &13.2 &32.6 & 33.3 & 12.1 &43.5  &42.2  & 28.3 &47.1  &28.2  \\
        Qwen3-vl-8B*~\cite{bai2025qwen3}  & 50 
        & 23.9  &34.8  &17.0 & 31.8 & 56.5 & 9.2 &33.3  &66.7  & 40.4 &21.7  &25.8  \\
        \midrule
        Ours w/ UITARS-1.5-7B  & 50 
        & 32.8 & 57.7 & 8.5 &39.3  &41.3  &8.6  & 29.8 &50.0  &27.5  & 54.4 & 28.2 \\
        Ours w/ Qwen3-vl-8B  & 50 
        & 34.8  &69.6  &17.0 & 19.6 & 26.1 & 13.1 &62.5  &40.0  & 17.7 &43.5  &\textbf{28.4}  \\
        \bottomrule
    \end{tabular}
    \label{tab:osworld}
\end{table*}

We also observe that increasing the interaction budget from 15 steps to 50 steps brings clear improvements to DynamicUI, with average accuracy rising from 15.5\% to 20.8\%. This trend suggests that DynamicUI can more effectively exploit longer decision horizons, which is particularly important for tasks requiring multi-step reasoning over dynamic interface changes.

A category-wise breakdown shows that DynamicUI performs best on InterruptUI (31.3\%) and DynList (22.7\%), demonstrating advantages in handling blocking system states and dynamically updated item collections. In contrast, performance is lower on ContentTrig (19.2\%) and EphemRef (18.8\%), where success depends on recovering short-term interaction history or grounding fleeting semantic cues. This suggests that DynamicUI is more effective for explicit and structured dynamic changes, while temporally sparse and content-conditioned signals remain challenging.

% We evaluate DynamicUI on the OSWorld benchmark in Tab.~\ref{tab:osworld}. To demonstrate the generality of our framework, we instantiate DynamicUI with two base agents, UITARS-1.5-7B~\cite{qin2025ui} and Qwen3-vl-8B~\cite{bai2025qwen3}. With UITARS-1.5-7B as the base agent, DynamicUI improves the average accuracy from 27.2\% to 28.2\% under the 50-step setting, with notable gains on Gimp, OS, Thunderbird, VLC, and VS Code. When instantiated with Qwen3-vl-8B, our framework reaches 28.4\% average accuracy, exceeding the Qwen3-vl-8B baseline by 2.6 points. These results suggest that DynamicUI is not limited to a single agent architecture and can provide competitive gains on a broader benchmark.

Moreover, we evaluate DynamicUI on the OSWorld benchmark, as summarized in Table~\ref{tab:osworld}. To demonstrate the generality of our framework, we instantiate DynamicUI with two different base agents, UITARS-1.5-7B~\cite{qin2025ui} and Qwen3-vl-8B~\cite{bai2025qwen3}. With UITARS-1.5-7B as the base agent, DynamicUI improves the average accuracy from 27.2\% to 28.2\% under the 50-step setting, with particularly notable gains on Gimp, OS, Thunderbird, VLC, and VS Code. When instantiated with Qwen3-vl-8B, our framework reaches 28.4\% average accuracy, exceeding the corresponding Qwen3-vl-8B baseline by 2.6 points. These results suggest that DynamicUI is not limited to a single agent architecture and can provide competitive gains on a broader benchmark.

\subsection{Ablation Study}
We conduct ablation studies on DynamicGUIBench to validate the effectiveness of the dynamic perceiver, refinement strategy, and reflection module, as shown in Tab.~\ref{tab:aba}.

As shown in Tab.~\ref{tab:aba}, the Dynamic Perceiver and reflection module are the main contributors to the overall gains. Adding DP improves the accuracy from 15.1\% to 17.4\%, while further introducing reflection boosts it to 20.8\%. In contrast, the refinement strategy brings only marginal improvements, likely because it mainly improves step-level actions, whose effect may be diluted by the overall trajectory under long-horizon evaluation.

\begin{table}[t]
\centering
\small
\setlength{\tabcolsep}{4pt}
\caption{Comparison of uniform frame sampling baselines and our Dynamic Perceiver (DP) across four POMDP categories on DynamicGUIBench.}
\label{tab:dp}
\resizebox{\linewidth}{!}{
\begin{tabular}{lccccc|c}
\toprule
\textbf{Method} & \textbf{Frames} 
& \textbf{Interruptive UI  } 
& \textbf{Ephemeral Ref.  } 
& \textbf{DynList } 
& \textbf{ContentTrig }
& \textbf{Avg. }
% & \textbf{Avg. Steps } 
\\
\midrule

Qwen3-vl-8B~\cite{bai2025qwen3} & - & 25.0& 12.3 & 13.6 & 19.2 & 15.1 \\
\midrule
\multirow{2}{*}{Uniform} 
& 1   & 25.0 & 15.3 & 31.8 & 3.9 & 16.8  \\
& 3   & 25.0 & 17.0 & 13.6 & 11.5 & 16.4 \\
\midrule
Ours & DP & 43.8 & 20.0 & 22.7 & 15.4 & \textbf{22.1} \\
\bottomrule
\end{tabular}
}
\end{table}

We compare the Dynamic Perceiver (DP) with uniform frame sampling using 1 and 3 frames, since the average number of frames selected by DP is close to three. Tab.~\ref{tab:dp} shows that simply sampling more frames does not guarantee better performance: Uniform-3 is slightly worse than Uniform-1 on average (16.4\% vs.\ 16.8\%). This indicates that performance depends more on selecting informative observations than on increasing the number of frames. By explicitly retrieving task-relevant historical context, DP achieves the best average accuracy of 22.1\%. Its gains are most evident on Interruptive UI (43.8\%) and EphemRef (20.0\%), where critical evidence is often tied to inserted interface states. DP also remains competitive on DynList and ContentTrig, demonstrating the effectiveness of adaptive perception over uniform history sampling.

We further compare Qwen3-VL-8B and GPT-5.4-mini as the reflection module on DynamicGUIBench. Tab.~\ref{tab:reflection} shows that using the same model for both execution and reflection is suboptimal, as the two roles may become contradictory or redundant. Reflection may suggest terminating the task while the executor continues acting, or simply duplicate reasoning already available to the executor. In contrast, GPT-5.4-mini provides more reliable high-level guidance and more accurate intermediate to-do lists, yielding consistent gains across all four categories and substantially better overall accuracy.

\begin{table}[t]
\centering
\small
\setlength{\tabcolsep}{4pt}
\caption{Comparison of different general-purpose models for the reflection module on DynamicGUIBench.}
\label{tab:reflection}
\resizebox{\linewidth}{!}{
\begin{tabular}{lcccc|c}
\toprule
\textbf{Method} 
& \textbf{Interruptive UI  } 
& \textbf{Ephemeral Ref.  } 
& \textbf{DynList } 
& \textbf{ContentTrig }
& \textbf{Avg. }
\\
\midrule
Qwen3-vl-8B~\cite{bai2025qwen3}  & 25.0& 9.3 & 9.1 & 11.5 & 11.3 \\
GPT-5.4-mini~\cite{singh2025openai} & 43.8 & 20.0 & 22.7 & 15.4 & \textbf{22.1} \\
\bottomrule
\end{tabular}
}
\end{table}

\subsection{Qualitative Success and Failure Analysis}
Fig.~\ref{fig:success_failure} presents a representative successful case of DynamicUI and a representative failure case of traditional GUI agents on DynamicGUIBench. In the top example, the environment undergoes several task-relevant interstitial changes during the web interaction. By explicitly tracking these transient states and incorporating them into the historical context, DynamicUI is able to recover the critical prompt \textit{`Cache Error, then click Refresh.'} and take the correct follow-up action, ultimately completing the task successfully. In contrast, the bottom example shows that traditional GUI agents rely primarily on the current observation and fail to recover this missing intermediate evidence once it disappears. As a result, they incorrectly interpret the page state, repeatedly click the `Login' button, and fall into a failure loop. This comparison highlights that, in dynamic GUI environments, successful decision-making often depends not only on the current screenshot but also on preserving and reasoning over task-critical interstitial states.

\section{Conclusion}
In this paper, we introduce DynamicGUIBench, the first partially observable benchmark designed to evaluate GUI agents under dynamic environments, covering four common dynamic processes in desktop environments. 
Our experiments show that current VLM-based GUI agents still struggle in such settings, highlighting a clear gap between existing capabilities and real-world GUI interaction demands. To address this challenge, we further propose DynamicUI, a framework consisting of three components: a dynamic perceiver, a reflection module, and a refinement strategy. Extensive results demonstrate that DynamicUI effectively improves performance in dynamic GUI environments while maintaining competitive performance on an additional public benchmark.

% \vspace{-2cm}
% \begin{acks}
% To Robert, for the bagels and explaining CMYK and color spaces.
% \end{acks}

\bibliographystyle{ACM-Reference-Format}
\bibliography{The_Name_of_the_Title_Is_Hope__1_/samples/sample-base}

\end{document}